\documentclass[letterpaper, 10 pt, conference]{ieeeconf}
\IEEEoverridecommandlockouts
\usepackage{multirow}
\usepackage[space,compress,sort]{cite}
\usepackage{amssymb}
\usepackage{amsmath}
\usepackage{bm}
\usepackage{booktabs}
\usepackage{graphicx}
\usepackage{xcolor}
\usepackage{cuted}    
\usepackage{capt-of}
\usepackage[linkcolor=black,citecolor=black,urlcolor=black,colorlinks=true]{hyperref}
\IEEEaftertitletext{\vspace{-2\baselineskip}}   
\title{\LARGE \bf Onboard Lightweight LLM Reasoning for Human-Aware Safe UAV Navigation}

\author{Siwei Cai$^{1}$, Yuwei Wu$^{2}$, Vijay Kumar$^{2}$, and Lifeng Zhou$^{1\dagger}$%
\thanks{$^{1}$Department of Electrical and Computer Engineering, Drexel University, USA.
        {\tt\small \{sc3568, lz457\}@drexel.edu}}%
\thanks{$^{2}$GRASP Lab, University of Pennsylvania, USA.
        {\tt\small \{yuweiwu, kumar\}@seas.upenn.edu}}%
\thanks{$^{\dagger}$Corresponding author.}%
}

\begin{document}
\maketitle
\thispagestyle{empty}
\pagestyle{empty}

\begin{strip}
\centering
\includegraphics[width=0.98\textwidth]{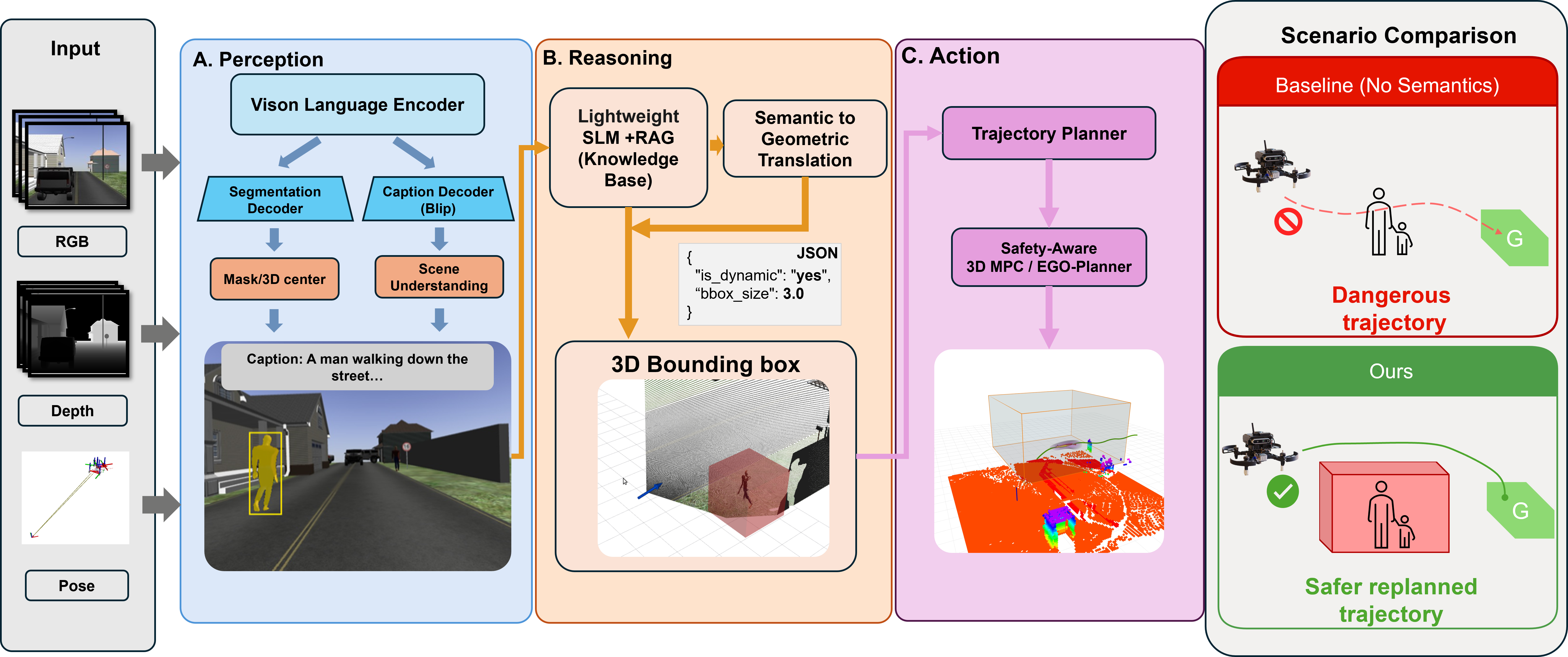}
\captionof{figure}{Overview. (A) Perception: a shared BLIP encoder feeds a captioning branch and a text-conditioned segmentation head; masks are fused with depth into 3D centroids. (B) Reasoning: a class table answers known categories instantly, while a RAG-augmented, grammar-constrained lightweight LLM handles unknown categories and outputs \texttt{\{is\_dynamic, bbox\_size\}}. (C) Action: obstacles receive persistent identities and are injected into the planner as unsafe boxes that are held through occlusions.}
\label{fig:TopOverview}
\end{strip}

\begin{abstract}
Aerial robots operating near people must treat a pedestrian, a parked car, and a wall differently, yet conventional perception--planning stacks reduce all of them to occupied geometry---a \emph{human-aware safety gap} in which category-specific clearance is unavailable to the planner and temporary occlusions remove pedestrian-specific constraints from its map.
We present a modular, fully onboard pipeline that closes this gap by turning visual observations into context-dependent safety constraints for real-time UAV navigation.
A shared BLIP encoder drives scene captioning and text-conditioned segmentation, which is fused with depth to localize obstacles in 3D.
Obstacle-specific risk is inferred by a lightweight LLM (Qwen-2.5-0.5B) stabilized with retrieval-augmented generation and grammar-constrained decoding that restricts its output to a planner-compatible JSON schema.
The LLM reasons once per object category rather than per frame: its verdicts are cached per class, so previously seen categories are resolved without inference and new ones trigger a fresh retrieval-grounded query; detections are additionally given persistent identities and held in the map across occlusions.
The perception pipeline runs at 10\,Hz on a Jetson Orin NX, while fallback LLM queries take approximately 0.8\,s and remain off the control-critical path.
In Gazebo, low-altitude urban navigation among pedestrians and vehicles shows a roughly $3\times$ reduction in collision events over a geometric planner, and semantics-aware landing among moving actors raises success from 34\% to 96\%.
The 224-pixel encoder input limits reliable small-object perception at range; conservative pedestrian margins and short-term persistence mitigate this limitation during navigation.

\end{abstract}

\section{Introduction}
\label{sec:introduction}
\vspace{-2mm}
UAVs are increasingly deployed for delivery, inspection, and monitoring in environments shared with people~\cite{saunders2024autonomous, aela2024uav, fang2024strategies}.
Such environments are not only dynamic but \emph{semantically heterogeneous}: a pedestrian, a moving truck, a parked car, and a hedge all constrain the feasible trajectory, yet they imply very different risk and should not receive identical avoidance behavior.
Conventional pipelines optimize trajectories well when the environment is accurately modeled~\cite{Hanover2021PerformancePAA, liu2017ral}, but they treat obstacles as geometry: occupancy maps and detections say \emph{where} an object is, not \emph{how} it should modify clearance, priority, or motion response~\cite{10802806}.

We call the resulting deficiency the \textbf{human-aware safety gap}: a geometric planner assigns a person and a lamp post the same clearance, and its map carries no notion of who or what is being avoided.
Human-aware planning should instead maintain larger, category-dependent clearance around people and preserve short-term awareness of them through temporary occlusions.
Our goal is therefore not to model people merely as dynamic obstacles, but to preserve category-specific safety semantics when planning around them: closing the gap means producing \emph{context-dependent} safety constraints---clearance, dynamism, and persistence that depend on the object's category and the operating context---and delivering them to the planner in real time, onboard.

Large language models (LLMs) offer the abstraction needed to produce such constraints~\cite{li2025large, Yue2025ASOA}, and LLM-based monitoring and planning has begun to appear in robotics~\cite{sinha2024real, wu2024hierarchical, wang2025contact}.
Sinha et al.~\cite{sinha2024real} pair a fast embedding classifier with a slow generative fallback, but the vehicle must wait while the large model reasons; contact-aware planners~\cite{wang2025contact} model interaction physics rather than grounding live observations.
Prior landing methods~\cite{niu2021vision, zhang2023coni} focus on cooperative targets and typically assume obstacle-free landing zones.
We ask instead: \textbf{\textit{how can language-based reasoning run \emph{onboard} an aerial robot so that its understanding of the people and vehicles around it becomes spatially grounded, real-time planner constraints?}}

Three system-level obstacles stand in the way.
(i) LLMs cannot consume raw imagery or 3D geometry, and end-to-end vision-language models (VLMs) are too heavy for Size, Weight and Power (SWaP)-limited platforms~\cite{bai2025qwen25vltechnicalreport}; we therefore decouple a light vision-language encoder (VLE)~\cite{li2022blip} from a text-only LLM and ground its output with depth.
(ii) Lightweight LLMs that fit onboard hallucinate~\cite{Sun2025OnionEvalAUA}; we stabilize them with retrieval-augmented generation (RAG)~\cite{Lewis2020RetrievalAugmentedGFA} over a safety knowledge base and constrain decoding to a JSON grammar.
(iii) Semantic guidance must not break real-time control; the LLM never commands the vehicle but adjusts planner-level constraints (unsafe regions and margins) consumed by a model-based planner and controller.
A fourth obstacle emerged only after closed-loop deployment among pedestrians: with a 2\,s map decay, a brief occlusion removes a person's constraint from the map before they reappear, so the pipeline needs \emph{object persistence} in addition to semantics.

This paper contributes:
(1) a semantic-to-spatial grounding pipeline that turns RGB-D observations into obstacle-specific risk and 3D safety constraints;
(2) a RAG-stabilized, grammar-constrained lightweight LLM reasoning module, whose per-class verdicts are cached so that the LLM reasons once per category rather than per frame, with a cache-miss reasoning pathway for categories outside a fixed semantic policy;
(3) persistent obstacle identities with occlusion-aware holding in the planner map; and
(4) an onboard feasibility study on Jetson-class hardware together with closed-loop Gazebo experiments in urban navigation among pedestrians and vehicles and in autonomous landing, with an analysis of perception under the limited input resolution that onboard compute imposes.

\vspace{-2mm}
\section{Method}
\label{sec:methodology}
Fig.~\ref{fig:TopOverview} shows the three subsystems. \emph{Perception} converts RGB-D frames into semantic and spatial primitives; \emph{Reasoning} maps them to obstacle-specific safety parameters; \emph{Action} keeps a persistent obstacle set and injects it into a model-based planner.

\subsection{Perception: Shared VLE, Segmentation, and 3D Grounding}
\label{sec:scene-understanding}
A single frozen BLIP ViT-B/16 encoder~\cite{li2022blip} ($\approx$86M parameters) serves two heads, avoiding redundant visual computation onboard.
The \emph{captioning branch} is BLIP's original decoder and yields a short zero-shot scene description (${\approx}30$ tokens) that identifies potential hazards.
The \emph{segmentation head} is a lightweight decoder we train on top of the encoder: multi-scale features from layers $L$, $L/2$, and $L/4$ are projected to a common width with $1{\times}1$ convolutions and fused via skip connections~\cite{ronneberger2015u}; a text prompt (e.g., ``person'', ``car'') conditions the fused map $\mathbf{F}_{\mathrm{vis}}$ through FiLM~\cite{perez2018film},
\begin{equation}
    \mathbf{F}_{\mathrm{cond}} = \mathbf{F}_{\mathrm{vis}} \odot \big(1 + \gamma(\mathbf{e}_{\mathrm{text}})\big) + \beta(\mathbf{e}_{\mathrm{text}}),
\end{equation}
before four resize-convolution up-blocks (bilinear upsampling followed by a $3{\times}3$ convolution, which avoids checkerboard artifacts~\cite{odena2016deconvolution}) produce the mask $\mathcal{M}$.
Each connected component of $\mathcal{M}$ that exceeds an area threshold is back-projected through the time-aligned depth map with intrinsics $\mathbf{K}$,
\begin{equation}
\label{equ:Pcenter}
    \mathbf{P}_{\mathrm{center}} = \tfrac{1}{|\mathcal{M}|} \textstyle\sum_{(u,v)\in\mathcal{M}} Z_{u,v}\,\mathbf{K}^{-1}[u,\,v,\,1]^{\top},
\end{equation}
after rejecting depths more than $2\sigma$ from the in-mask median.
The result is a list of $(\text{class}, \mathbf{P}_{\mathrm{center}}, \text{confidence})$ tuples at 10\,Hz.

\subsection{Reasoning: From Semantics to Safety Parameters}
\label{sec:semantic-reasoning}
\textbf{LLM-populated verdict cache.}
The LLM does not run per frame; it runs \emph{per class}.
The first time a class label appears, its name and the scene caption are sent through retrieval and the LLM, which returns a semantic verdict $\{\texttt{is\_dynamic}, \texttt{bbox\_size}\}$; the verdict is stored in a per-class cache and reused for every later detection of that class at no inference cost.
The cache is therefore not a hand-written lookup table but the accumulated output of the reasoning module: it is warm-started with the verdicts the same RAG--LLM path produced for the categories the segmentation head is prompted with (person, car, truck), and it grows whenever a new label is encountered.
This mechanism is designed to extend the pipeline beyond a fixed semantic policy: labels absent from the cache can trigger retrieval-grounded reasoning and subsequently be handled like cached categories. In our 20-run urban benchmark, however, every observed label hit the warm cache, so no online LLM query was issued after initialization; the landing ablation in Sec. IV-B isolates the contribution of the reasoning path itself.
In our 20-run urban benchmark every label hit the warm cache, so no online LLM query was issued after initialization; the landing ablation in Sec.~\ref{sec:landing} isolates the contribution of the reasoning path itself.
Because a novel object can appear at any moment, the full stack is loaded at start-up rather than lazily, and a verdict that fails grammar validation falls back to a conservative default box rather than discarding the object.

\textbf{Hybrid retrieval.}
For an unknown class, the caption and class name query a local safety knowledge base (KB) with a hybrid search: dense retrieval (\texttt{bge-small-en-v1.5}, ${<}100$\,MB) captures synonyms (``pedestrian''$\rightarrow$``person''), while sparse BM25 anchors high-priority keywords (``truck'').
Scores are fused with Reciprocal Rank Fusion~\cite{cormack2009reciprocal}, $S(d)=\sum_{m}1/(\eta+r_m(d))$, $\eta{=}60$, and only the top-$k$ segments enter the prompt to avoid attention dilution in small models~\cite{fang2025attentionrag}.

\textbf{Edge inference and grammar-constrained output.}
Reasoning runs Qwen-2.5-0.5B-Instruct on \texttt{llama.cpp} with importance-matrix 4-bit quantization~\cite{frantar2022gptq} and a statically allocated KV cache for constant memory and predictable latency.
A GBNF grammar restricts decoding to the expected \texttt{\{"is\_dynamic": yes|no, "bbox\_size": float\}} schema, which substantially reduces malformed generations and removes the need for a separate free-form parsing stage.
Following \emph{Code as Policies}~\cite{liang2022code}, the LLM is deliberately restricted to \emph{semantic} risk (dynamic or not, how much clearance) and does not produce waypoints: small models are poor at Euclidean arithmetic~\cite{bubeck2023sparks}, and encoding dense geometry into the prompt inflates context and degrades reasoning~\cite{liu2024lost}.

\subsection{Action: Persistent Obstacles and Planner Constraints}
\label{sec:action}
\textbf{Persistent identity and occlusion hold.}
A per-frame detection list is not enough for planning near people: temporary occlusions (e.g., a pedestrian passing behind a parked vehicle) create gaps in detections, and with a fixed voxel decay the corresponding constraint leaves the map within seconds, motivating short-term persistence in the planner map.
We assign each detection a stable object id by greedy nearest-neighbor association within its class (gate $2$\,m, time-to-live $4$\,s) and publish the semantic box together with that id.
The planner map keeps, per id, the last-seen time and last known box; an object whose age is below a hold time $T_h$ (3\,s) stays in the map at its last position, and only afterwards are its voxels released.
Holding in place rather than extrapolating is deliberate for pedestrians, whose heading is not persistent; vehicles can be extrapolated with a tracked velocity when available. Human awareness thus enters the planner through differentiated clearance and persistence policies rather than by treating people as generic occupied space.
Ids also expose a metric we use in Sec.~\ref{sec:results}: how many objects are carried across detection gaps.

\textbf{Unsafe regions in the planner.}
Every dynamic object yields an axis-aligned box $\mathcal{B}_i$ of side \texttt{bbox\_size} around $\mathbf{P}_{\mathrm{center}}$; the unsafe set is $\mathcal{B}=\bigcup_i\mathcal{B}_i$.
The planner interface is intentionally minimal so that the back end can be swapped.
For the landing case study we follow~\cite{9531427}: a kinodynamic A* front end whose heuristic is augmented with a soft penalty $\tilde h(\mathbf{p}) = h_0(\mathbf{p}) + \lambda\,\phi(\mathbf{p})$, where $\phi$ grows quadratically with penetration into $\mathcal{B}$ and states inside $\mathcal{B}$ are pruned, followed by a nonlinear MPC whose safe-flight corridor $\tilde{\mathcal{P}}_c\subseteq\mathcal{P}_c$ is regenerated with $\tilde{\mathcal{P}}_c\cap\mathcal{B}=\emptyset$, and an $SE(3)$ geometric controller~\cite{5717652} for tracking.
For the urban navigation experiments the same boxes are injected as inflated voxels into the ESDF-free grid map of a B-spline local planner (EGO-Planner~\cite{zhou2021ego}), demonstrating that the semantic layer is planner-agnostic.

\vspace{-2mm}
\section{Onboard Feasibility}
\label{sec:feasibility}
\textbf{Why not the cloud.}
Field radio telemetry gives us ${\approx}100$\,kbps uplink, so a single $384{\times}384$ JPEG (${\approx}800$\,kbit) costs ${\approx}8$\,s to transmit; measured JSON-generation latency of hosted models adds $1.6$--$2.8$\,s on average with spikes above $4.7$\,s (GPT-3.5-Turbo/GPT-4o/GPT-5.1-low: 1.62/2.80/2.42\,s).
A roughly $10$\,s round trip could correspond to about 10\,m of travel at $1$\,m/s before a semantic result is available. Everything therefore runs onboard.

\textbf{Platform.}
We compared an Intel NUC~13 Pro (CPU only, ${>}40$\,W, 256\,g decased), a Jetson Orin Nano (8\,GB, 15\,W), and a Jetson Orin NX (16\,GB, 25\,W, 100 TOPS, ${\approx}180$\,g with carrier).
The decased NUC weighs 256\,g, compared with approximately 180\,g for the Orin NX with carrier, draws more than 40\,W, and lacks tensor cores; in our implementation the 8\,GB Orin Nano could not accommodate the VLE, KB, LLM, VIO, and ROS stack simultaneously. We deploy on the Orin NX.

\textbf{Modular VLE+LLM versus monolithic small VLMs.}
Monolithic small VLMs project an image into 144--576 visual tokens that must traverse the full language backbone, and rely on 300--675M-parameter encoders.
Our ``compress-then-reason'' design sends ${\approx}30$ text tokens through a 0.5B model behind an 86M encoder, reducing the number of language-model input tokens by more than $19\times$ relative to the compared monolithic VLMs (Table~\ref{tab:vlm_comparison}); the same encoder output is reused by the segmentation head.

\begin{table}[t]
\centering
\caption{Modular vs.\ monolithic pipelines on Jetson Orin NX (16\,GB).}
\label{tab:vlm_comparison}
\footnotesize
\setlength{\tabcolsep}{3pt}
\begin{tabular}{lccc}
\toprule
\textbf{Pipeline} & \textbf{Encoder} & \textbf{LLM tokens} & \textbf{Latency} \\
\midrule
Qwen2-VL-2B & SigLIP-style 675M & 144 & 2.77\,s \\
PaliGemma-3B & SigLIP 400M & 256 & 4.75\,s \\
SmolVLM & SigLIP 400M & $81{\times}$tiles & 7.54\,s \\
LLaVA-Phi-3 & CLIP-L 304M & 576 & 7.62\,s \\
\midrule
BLIP + Llama-3.2-1B & ViT-B 86M & ${\approx}30$ & 0.88\,s \\
\textbf{BLIP + Qwen-2.5-0.5B} & \textbf{ViT-B 86M} & \textbf{${\approx}30$} & \textbf{0.79\,s} \\
\bottomrule
\end{tabular}
\vspace{-0.3cm}
\end{table}

\textbf{LLM selection.}
We benchmarked Llama-3.1/3.2, Qwen-2.5, Gemma-2 and Phi-3.5 families at Q3--FP16 on the Orin NX for peak memory, generation time, and instruction following (valid JSON without filler).
Gemma-2, Phi-3.5, and aggressively quantized Llama-3.2-1B (IQ4, Q3) frequently failed to follow the required output schema under the tested quantization settings; Llama-3.1-8B fits in memory but exceeds $3.3$\,s per query.
Qwen-2.5-0.5B (Q4\_K\_M) answers in ${\approx}0.75$--$0.79$\,s using $5.6$\,GB of total system memory, leaving headroom for perception and VIO; on the desktop-class simulation host the same call takes $0.21$--$0.41$\,s.
With the per-class verdict cache of Sec.~\ref{sec:semantic-reasoning} this latency is incurred once per new category rather than per frame; the perception loop maintained $9.96$\,Hz over 20 runs.

\vspace{-2mm}
\section{Experiments}
\label{sec:results}
All closed-loop experiments use ROS Noetic with Gazebo physics and simulated RGB-D sensing~\cite{Furrer2016}.
The primary evaluation is low-altitude urban navigation among pedestrians and vehicles (Sec.~\ref{sec:urban}), the setting that motivated the human-aware safety gap and where the persistence mechanism was introduced (EGO-Planner back end); autonomous landing (Sec.~\ref{sec:landing}) serves as a second case study with the kinodynamic-A*/MPC back end.

\subsection{Urban Navigation Among Pedestrians and Vehicles}
\label{sec:urban}
\textbf{Setup.}
A town world contains two scripted SUVs (one crossing the UAV's path head-on, one emerging from behind a wall at 1.5--2\,m/s) and two walkers (one along the road at 1.2\,m/s, one performing a bounded random walk at 1\,m/s).
The UAV flies a fixed multi-goal route at low altitude for ${\approx}106$\,s; the perception front end runs the fixed prompt set \{person, car, truck\} at 10\,Hz.
We compare the geometric planner (\emph{off}) against the full semantic pipeline with persistent ids and $T_h{=}3$\,s (\emph{on}), 20 runs each with identical goals and spawn points.

\textbf{Closed-loop safety.}
Collision events per run drop from $1.35$ to $0.45$ ($p{=}0.038$), the only safety metric that reaches significance at $n{=}20$; worst-case clearance improves in mean ($0.16\rightarrow0.52$\,m) but not significantly, and crash counts (3 vs.\ 1 of 20) are within noise.
The semantic planner maintains larger human-aware separation through additional detours, at the cost of path length ($54.5\rightarrow68.7$\,m) and peak jerk ($2.6\times$).
The simulator is not deterministic; with a per-run worst-clearance standard deviation of $0.5$--$0.8$\,m, differences below ${\approx}0.2$\,m are not resolvable at $n{=}20$.

\textbf{Persistence and the verdict cache.}
Per run, an average of 45 object ids are created, 23\% of which die after a single frame (background artifacts); 23 survive at least $1$\,s, and \textbf{20 are carried across a detection gap of ${\geq}0.5$\,s}, with the longest bridged gap equal to the $4$\,s time-to-live and single ids surviving up to $70$\,s.
All observed classes were served from the warm verdict cache during these runs, so no online LLM query was issued, and perception maintained $9.96$\,Hz.

\begin{figure}[t]
    \centering
    \includegraphics[width=0.8\linewidth]{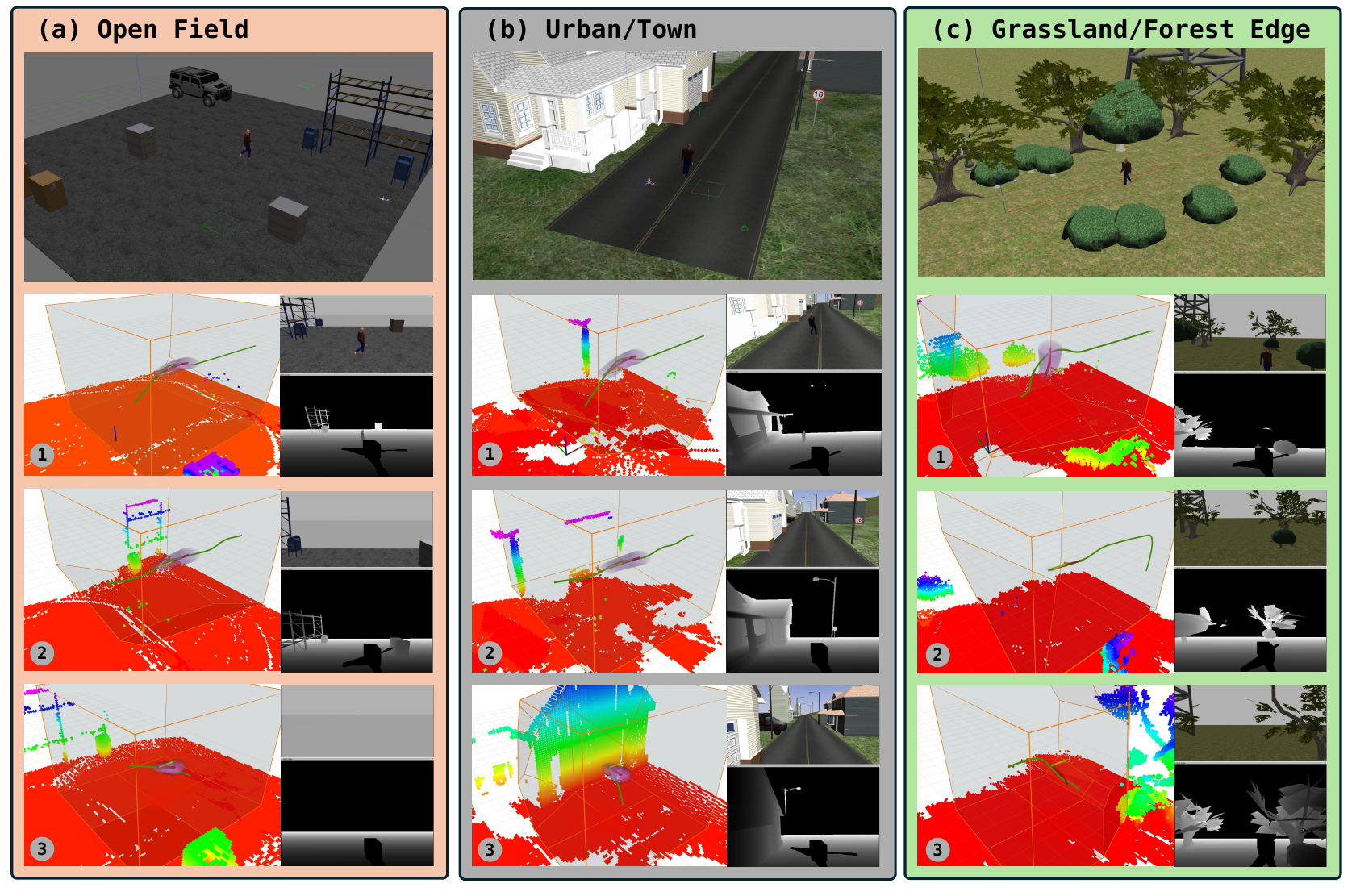}
    \caption{Landing environments: (a) open field with moving humanoid actors and a vehicle; (b) urban town with buildings, streets and parked cars; (c) grassland/forest edge with trees and underbrush.}
    \label{fig:testenv}
    \vspace{-0.4cm}
\end{figure}

\subsection{Case Study: Semantics-Aware Landing}
\label{sec:landing}
\textbf{Setup and baselines.}
Three environments (Fig.~\ref{fig:testenv}) test generalization; the open field additionally contains a humanoid walking unpredictably near the landing pad.
A trial succeeds if the UAV lands within 0.5\,m of the target without entering the actor's 1\,m safety buffer (50 trials per cell).
We compare (i) \emph{Baseline MPC}, the same planner with depth-only geometric avoidance; (ii) \emph{LLM--MPC}, the LLM acting on BLIP captions without RAG; and (iii) \emph{Ours}, the full RAG-stabilized pipeline.

\begin{table}[h]
  \centering
  \caption{Landing success rate (\%), 50 trials per entry.}
  \label{tab:landing}
  \small
  \begin{tabular}{@{}lccc@{}}
    \toprule
    \textbf{Method} & \textbf{Open field (dyn.)} & \textbf{Urban} & \textbf{Forest} \\
    \midrule
    Baseline MPC & 34.0 & 28.0 & 32.0 \\
    LLM--MPC (no RAG) & 6.0 & 12.0 & 6.0 \\
    \textbf{Ours} & \textbf{96.0} & \textbf{74.0} & \textbf{80.0} \\
    \bottomrule
  \end{tabular}
  \vspace{-0.2cm}
\end{table}

\textbf{Results.}
Table~\ref{tab:landing} shows that geometric avoidance alone reaches 34\% among moving actors because it cannot anticipate that a pedestrian requires a larger, earlier safety margin than a static object, while the unaugmented LLM is \emph{worse} than no LLM at all (6--12\%): hallucinated or malformed verdicts destabilize the planner.
RAG plus grammar-constrained decoding turns the same 0.5B model into a reliable source of safety constraints.

\subsection{Perception Under Limited Input Resolution}
\label{sec:limitations}
The 224-pixel encoder input limits the effective resolution available for small, distant objects.
This effect is most noticeable for pedestrians, which occupy a much smaller image area than vehicles at the same range.
The larger pedestrian margin and short-term persistence help maintain conservative safety behavior under this sensing limitation.
We therefore use a larger pedestrian safety margin and preserve pedestrian-specific constraints through short occlusions.
Higher-resolution perception within the same onboard compute budget is left for future work.

\section{Conclusion}
\label{sec:conclusion}
We showed that a 0.5B-parameter LLM, grounded by retrieval, constrained to a machine-parseable schema, kept off the per-frame critical path by per-class verdict caching, and complemented by object persistence in the planner map, can supply \emph{context-dependent safety constraints for aerial robots operating around people}: a pedestrian and a parked car receive different clearance, and pedestrian-specific constraints persist through short occlusions.
The full stack runs on a Jetson Orin NX, reduces collision events roughly threefold in urban navigation among pedestrians and vehicles---at the cost of longer paths and sharper maneuvers---and raises landing success among moving actors from 34\% to 96\%.
In the evaluated closed-loop scenarios, all observed categories were served from the warm verdict cache; evaluating retrieval-grounded reasoning for previously unseen categories, including comparison against fixed semantic policies, remains future work. Small-object perception at range is another remaining limitation, primarily due to the 224-pixel encoder input. Future work will therefore investigate higher-resolution perception within the same compute budget, improved per-class motion models for persistent objects, and cache-miss reasoning for novel categories.

\bibliography{reference}

@inproceedings{li2022blip,
  title={Blip: Bootstrapping language-image pre-training for unified vision-language understanding and generation},
  author={Li, Junnan and Li, Dongxu and Xiong, Caiming and Hoi, Steven},
  booktitle={International conference on machine learning},
  pages={12888--12900},
  year={2022},
  organization={PMLR}
}

@article{odena2016deconvolution,
  title={Deconvolution and checkerboard artifacts},
  author={Odena, Augustus and Dumoulin, Vincent and Olah, Chris},
  journal={Distill},
  volume={1},
  number={10},
  pages={e3},
  year={2016}
}

@inproceedings{perez2018film,
  title={Film: Visual reasoning with a general conditioning layer},
  author={Perez, Ethan and Strub, Florian and De Vries, Harm and Dumoulin, Vincent and Courville, Aaron},
  booktitle={Proceedings of the AAAI conference on artificial intelligence},
  volume={32},
  number={1},
  year={2018}
}

@inproceedings{ronneberger2015u,
  title={U-net: Convolutional networks for biomedical image segmentation},
  author={Ronneberger, Olaf and Fischer, Philipp and Brox, Thomas},
  booktitle={International Conference on Medical image computing and computer-assisted intervention},
  pages={234--241},
  year={2015},
  organization={Springer}
}

@article{liang2022code,
  title={Code as policies: Language model programs for embodied control},
  author={Liang, Jacky and Huang, Wenlong and Xia, Fei and Xu, Peng and Hausman, Karol and Ichter, Brian and Florence, Pete and Zeng, Andy},
  journal={arXiv preprint arXiv:2209.07753},
  year={2022}
}

@misc{llama.cpp,
  author = {Gerganov, Georgi},
  title = {{ggml-org/llama.cpp: Port of Facebook's LLaMA model in C/C++}},
  year = {n.d.},
  publisher = {GitHub},
  howpublished = {\url{https://github.com/ggml-org/llama.cpp}},
}

@misc{bai2025qwen25vltechnicalreport,
      title={Qwen2.5-VL Technical Report}, 
      author={Shuai Bai and et al.},
      year={2025},
      eprint={2502.13923},
      archivePrefix={arXiv},
      primaryClass={cs.CV},
      url={https://arxiv.org/abs/2502.13923}, 
}

@article{wu2024hierarchical,
  title={Hierarchical LLMs In-the-Loop Optimization for Real-Time Multi-Robot Target Tracking under Unknown Hazards},
  author={Wu, Yuwei and Tao, Yuezhan and Li, Peihan and Shi, Guangyao and Sukhatmem, Gaurav S and Kumar, Vijay and Zhou, Lifeng},
  journal={arXiv preprint arXiv:2409.12274},
  year={2024}
}

@article{sinha2024real,
  title={Real-time anomaly detection and reactive planning with large language models},
  author={Sinha, Rohan and Elhafsi, Amine and Agia, Christopher and Foutter, Matthew and Schmerling, Edward and Pavone, Marco},
  journal={arXiv preprint arXiv:2407.08735},
  year={2024}
}

@misc{zhang2023coni,
  title={CoNi-MPC: cooperative non-inertial frame based model predictive control. IEEE Robot Autom Lett 8 (12): 8082--8089},
  author={Zhang, B and others},
  year={2023}
}

@article{niu2021vision,
  title={Vision-based autonomous landing for unmanned aerial and ground vehicles cooperative systems},
  author={Niu, Guanchong and Yang, Qingkai and Gao, Yunfan and Pun, Man-On},
  journal={IEEE robotics and automation letters},
  volume={7},
  number={3},
  pages={6234--6241},
  year={2021},
  publisher={IEEE}
}

@article{frantar2022gptq,
  title={Gptq: Accurate post-training quantization for generative pre-trained transformers},
  author={Frantar, Elias and Ashkboos, Saleh and Hoefler, Torsten and Alistarh, Dan},
  journal={arXiv preprint arXiv:2210.17323},
  year={2022}
}

@article{wang2025contact,
  title={Contact-Aware Motion Planning Among Movable Objects},
  author={Wang, Haokun and Wang, Qianhao and Gao, Fei and Shen, Shaojie},
  journal={arXiv preprint arXiv:2502.03317},
  year={2025}
}

@article{fang2024strategies,
  title={Strategies for optimized uav surveillance in various tasks and scenarios: A review},
  author={Fang, Zixuan and Savkin, Andrey V},
  journal={Drones},
  volume={8},
  number={5},
  pages={193},
  year={2024},
  publisher={MDPI}
}

@article{aela2024uav,
  title={UAV-based studies in railway infrastructure monitoring},
  author={Aela, Peyman and Chi, Hung-Lin and Fares, Ali and Zayed, Tarek and Kim, Minkoo},
  journal={Automation in Construction},
  volume={167},
  pages={105714},
  year={2024},
  publisher={Elsevier}
}

@article{saunders2024autonomous,
  title={Autonomous aerial robotics for package delivery: A technical review},
  author={Saunders, Jack and Saeedi, Sajad and Li, Wenbin},
  journal={Journal of Field Robotics},
  volume={41},
  number={1},
  pages={3--49},
  year={2024},
  publisher={Wiley Online Library}
}

@article{liu2024lost,
  title={Lost in the middle: How language models use long contexts},
  author={Liu, Nelson F and Lin, Kevin and Hewitt, John and Paranjape, Ashwin and Bevilacqua, Michele and Petroni, Fabio and Liang, Percy},
  journal={Transactions of the Association for Computational Linguistics},
  volume={12},
  pages={157--173},
  year={2024}
}

@article{bubeck2023sparks,
  title={Sparks of artificial general intelligence: early experiments with GPT-4 (2023)},
  author={Bubeck, S{\'e}bastien and Chandrasekaran, Varun and Eldan, Ronen and Gehrke, Johannes and Horvitz, Eric and Kamar, Ece and Lee, Peter and Lee, Yin Tat and Li, Yuanzhi and Lundberg, Scott and others},
  journal={arXiv preprint arXiv:2303.12712},
  volume={1},
  year={2023}
}

@article{Hanover2021PerformancePAA,
  title={Performance, Precision, and Payloads: Adaptive Nonlinear MPC for Quadrotors},
  author={D. Hanover and Philipp Foehn and Sihao Sun and Elia Kaufmann and D. Scaramuzza},
  journal={IEEE Robotics and Automation Letters},
  year={2021},
  volume={7},
  pages={690-697},
  url={http://ieeexplore.ieee.org/stamp/stamp.jsp?tp=\&arnumber=9632352}
}

@article{li2025large,
  title={Large Language Models for Multi-Robot Systems: A Survey},
  author={Li, Peihan and An, Zijian and Abrar, Shams and Zhou, Lifeng},
  journal={arXiv preprint arXiv:2502.03814},
  year={2025}
}

@article{Yue2025ASOA,
  title={A Survey of Large Language Model Agents for Question Answering},
  author={Murong Yue},
  journal={ArXiv},
  year={2025},
  volume={abs/2503.19213},
  url={https://api.semanticscholar.org/CorpusId:277313791}
}

@article{Sun2025OnionEvalAUA,
  title={OnionEval: An Unified Evaluation of Fact-conflicting Hallucination for Small-Large Language Models},
  author={Chongren Sun and Yuran Li and Di Wu and Benoit Boulet},
  journal={ArXiv},
  year={2025},
  volume={abs/2501.12975},
  url={https://arxiv.org/pdf/2501.12975.pdf}
}

@inproceedings{cormack2009reciprocal,
  title={Reciprocal rank fusion outperforms condorcet and individual rank learning methods},
  author={Cormack, Gordon V and Clarke, Charles LA and Buettcher, Stefan},
  booktitle={Proceedings of the 32nd international ACM SIGIR conference on Research and development in information retrieval},
  pages={758--759},
  year={2009}
}

@article{fang2025attentionrag,
  title={Attentionrag: Attention-guided context pruning in retrieval-augmented generation},
  author={Fang, Yixiong and Sun, Tianran and Shi, Yuling and Gu, Xiaodong},
  journal={arXiv preprint arXiv:2503.10720},
  year={2025}
}

@article{Lewis2020RetrievalAugmentedGFA,
  title={Retrieval-augmented generation for knowledge-intensive nlp tasks},
  author={Lewis, Patrick and Perez, Ethan and Piktus, Aleksandra and Petroni, Fabio and Karpukhin, Vladimir and Goyal, Naman and K{\"u}ttler, Heinrich and Lewis, Mike and Yih, Wen-tau and Rockt{\"a}schel, Tim and others},
  journal={Advances in neural information processing systems},
  volume={33},
  pages={9459--9474},
  year={2020}
}

@INPROCEEDINGS{5717652,
  author={Lee, Taeyoung and Leok, Melvin and McClamroch, N. Harris},
  booktitle={49th IEEE Conference on Decision and Control (CDC)}, 
  title={Geometric tracking control of a quadrotor UAV on SE(3)}, 
  year={2010},
  volume={},
  number={},
  pages={5420-5425},
  doi={10.1109/CDC.2010.5717652}}

@article{liu2017ral,
  title={Planning Dynamically Feasible Trajectories for Quadrotors using Safe Flight Corridors in 3-D Complex Environments},
  author={Liu, Sikang and Watterson, Michael and Mohta, Kartik and Sun, Ke and Bhattacharya, Subhrajit and Taylor, Camillo Jose and Kumar, Vijay},
  journal={IEEE Robotics and Automation Letters ({RA-L})},
  pages={1688--1695},
  year={2017},
  publisher={IEEE}
}

@INPROCEEDINGS{10802806,
  author={Wang, Dong and Wang, Jingping and He, Suqin and Huang, Jinxin and Zhang, Bangyan and Mao, Yinian and Huang, Guoquan and Xu, Chao and Gao, Fei},
  booktitle={2024 IEEE/RSJ International Conference on Intelligent Robots and Systems (IROS)}, 
  title={Multi-Fov-Constrained Trajectory Planning for Multirotor Safe Landing}, 
  year={2024},
  volume={},
  number={},
  pages={5356-5363},
  doi={10.1109/IROS58592.2024.10802806}}

@ARTICLE{9531427,
  author={Wu, Yuwei and Ding, Ziming and Xu, Chao and Gao, Fei},
  journal={IEEE Robotics and Automation Letters}, 
  title={External Forces Resilient Safe Motion Planning for Quadrotor}, 
  year={2021},
  volume={6},
  number={4},
  pages={8506-8513},
  doi={10.1109/LRA.2021.3110316}}

@Inbook{Furrer2016,
author="Furrer, Fadri
and Burri, Michael
and Achtelik, Markus
and Siegwart, Roland",
editor="Koubaa, Anis",
chapter="RotorS---A Modular Gazebo MAV Simulator Framework",
title="Robot Operating System (ROS): The Complete Reference (Volume 1)",
year="2016",
publisher="Springer International Publishing",
address="Cham",
pages="595--625",
isbn="978-3-319-26054-9",
doi="10.1007/978-3-319-26054-9_23",
url="http://dx.doi.org/10.1007/978-3-319-26054-9_23"
}

@article{zhou2021ego,
  title={{EGO-Planner}: An {ESDF}-free gradient-based local planner for quadrotors},
  author={Zhou, Xin and Wang, Zhepei and Ye, Hongkai and Xu, Chao and Gao, Fei},
  journal={IEEE Robotics and Automation Letters},
  volume={6},
  number={2},
  pages={478--485},
  year={2021}
}

\end{document}